\documentclass[letterpaper]{article} 
\usepackage{aaai23}  
\usepackage{times}  
\usepackage{helvet}  
\usepackage{courier}  
\usepackage[hyphens]{url}  
\usepackage{graphicx} 
\def\UrlFont{\rm}  
\usepackage{natbib}  
\usepackage{caption} 
\usepackage{algorithm}
\usepackage{algorithmic}
\usepackage{amssymb}
\usepackage{amsmath,amsfonts}
\usepackage{multirow}
\usepackage{booktabs}

\usepackage{newfloat}
\usepackage{listings}
\DeclareCaptionStyle{ruled}{labelfont=normalfont,labelsep=colon,strut=off} 
\floatstyle{ruled}
\newfloat{listing}{tb}{lst}{}
\floatname{listing}{Listing}
\begin{document}

\title{TOT: Topology-Aware Optimal Transport for Multimodal Hate Detection}
\author{
    Linhao Zhang\textsuperscript{\rm 1,2,3},
    Li Jin\textsuperscript{\rm 1,2}\thanks{Corresponding Author.},
    Xian Sun\textsuperscript{\rm 1,2}, 
    Guangluan Xu\textsuperscript{\rm 1,2}, Zequn Zhang\textsuperscript{\rm 1,2}, \\Xiaoyu Li\textsuperscript{\rm 1,2},
    Nayu Liu\textsuperscript{\rm 1,2,3},
    Qing Liu\textsuperscript{\rm 1,2},
    Shiyao Yan\textsuperscript{\rm 1,2,3}
}
\affiliations{
    \textsuperscript{\rm 1}Aerospace Information Research Institute, Chinese Academy of Sciences \\
    
    \textsuperscript{\rm 2}Key Laboratory of Network Information System Technology (NIST), Aerospace Information Research Institute
   
    \textsuperscript{\rm 3}School of Electronic, Electrical and Communication Engineering, University of Chinese Academy of Sciences\\
    
   zhanglinhao20@mails.ucas.ac.cn, jinlimails@gmail.com\\
}

\maketitle

\begin{abstract}
\begin{quote}

Multimodal hate detection, which aims to identify the harmful content online such as memes, is crucial for building a wholesome internet environment. Previous work has made enlightening exploration in detecting explicit hate remarks. However, most of their approaches neglect the analysis of implicit harm, which is particularly challenging as explicit text markers and demographic visual cues are often twisted or missing. The leveraged cross-modal attention mechanisms also suffer from the distributional modality gap and lack logical interpretability. To address these 
semantic gap issues, we propose TOT: a topology-aware optimal transport framework to decipher the implicit harm in memes scenario, which formulates the cross-modal aligning problem as solutions for optimal transportation plans. Specifically, we leverage an optimal transport kernel method to capture complementary information from multiple modalities. The kernel embedding provides a non-linear transformation ability to reproduce a kernel Hilbert space (RKHS), which reflects significance for eliminating the distributional modality gap. Moreover, we perceive the topology information based on aligned representations to conduct bipartite graph path reasoning. The newly achieved state-of-the-art performance on two publicly available benchmark datasets, together with  further visual analysis, demonstrate the superiority of TOT in capturing implicit cross-modal alignment.

\end{quote}
\end{abstract}

\section{Introduction}


\noindent The flourishing expansion of social media has facilitated
the exchange of opinions among individuals, different cultural or social communities. However, under the influence of major events such as the Russian-Ukrainian conflict and COVID-19, these platforms are flooded with hateful content, disseminating discriminatory statements toward social groups based
on their races, religions or other characteristics. Such hateful content is sowing seeds of disunity to exacerbate violent and
criminal behaviors in  conflict areas. Therefore, the automatic identification of hateful content on the Internet is a social research issue with great urgency.

\begin{figure}[!htbp]
    \centering
    \includegraphics[width=0.48\textwidth]{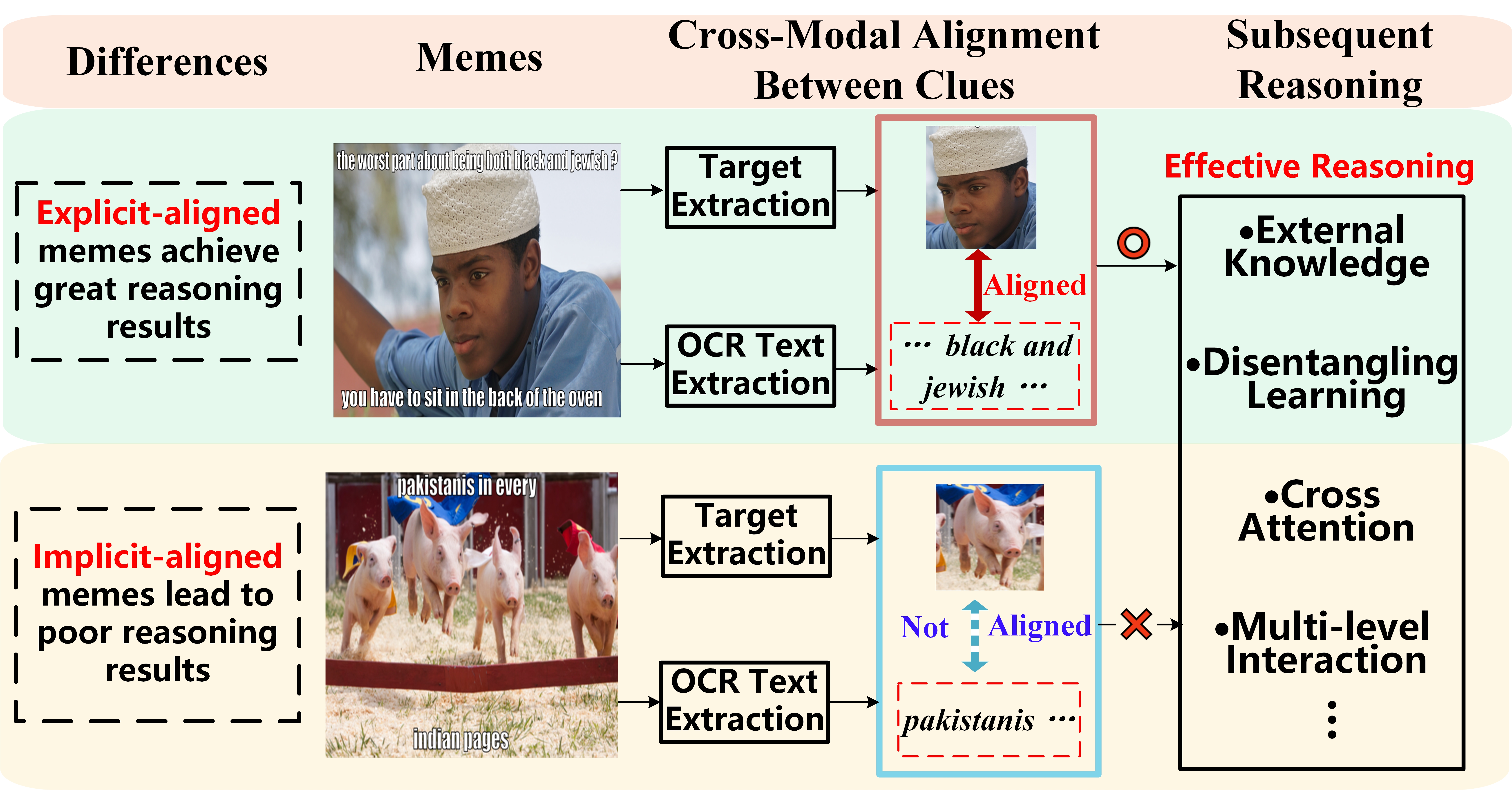}
    \caption{Differences between explicit-aligned and implicit-aligned memes. The aligning procedure is naturally completed in explicit harmful content, but is indispensable for implicit cases.}
    \label{intro}
\end{figure}
Memes, as a popular kind of social media user-generated content (UGC), have become a prevalent method for propagating harmful content due to their viral nature. Typically, a meme is an image embedded with a short piece of text. Although memes have a puckish sense of humor, some implicit hateful content is concealed. The critical factor in detecting harmful memes is combining well-aligned visual-linguistic clues and conducting multimodal reasoning.

Previous works have made remarkable progress in detecting explicit-aligned hate.  \cite{DBLP:conf/mm/LeeCFJC21,DBLP:conf/emnlp/PramanickSDAN021}. Their models achieved various kinds of practical reasoning through external knowledge enhancement  \cite{zhu2020enhance,zhou2021multimodal}, sophisticated interaction mechanisms \cite{lippe2020multimodal} or an  ensemble of multiple visual-linguistic models \cite{das2020detecting,velioglu2020detecting}. However, these approaches neglect the nuance of the multimodal aligning process, which is necessary for subsequent reasoning procedures. Take the second meme in Figure \ref{intro} as an example, which dehumanizes Pakistanis as pigs. Although the word 'Pakistanis' has no semantic correspondence with the 'pigs' in the picture, a human can easily capture the implicit cross-modal alignment between 'Pakistanis' and 'pigs' to perceive the potent offense. Nevertheless, this implicit harm is challenging for 
efficient detection. Previous works neglected the aligning procedure and conducted the reasoning process directly \cite{PanL0Q020}, resulting in suboptimal performance on implicit-aligned memes.

Recently, optimal transport (OT), as one of the research hotspots from optimization theory, has attracted extensive attention in computer vision, natural language processing and other fields due to its excellent performance on sequence alignment and domain adaption problems \cite{DuanCTYXZC22}. OT has the ability to reduce distributional bias in a more interpretable way under low-resource scenarios, which solves the problem of lacking external labels for aligning multimodal clues in implicit harmful memes \cite{ChenG0LC020, MareticGCF22}.



In this paper, inspired by the OT theory, we propose a general framework TOT: topology-aware optimal transport to capture implicit cross-modal alignment in memes. Specifically, we leverage an optimal transport kernel method to reformulate the alignment problem across different modalities. The leveraged Gaussian kernel provides the transformation ability to reproduce kernel Hilbert space (RKHS), in which the distributional modality gap gets eliminated to calculate an informative cost matrix for generating transportation plans. Secondly, based on the sinkhorn algorithm, we acquire optimal transportation plans, which are used for assigning source values to target distribution at minimum total cost. In our implementation, the cost matrix is a divergence of alignment represented by the pair-wise dot product of image and text sequence features in RHKS. Thus the optimal transportation plan with minimum total cost represents the assigning weights with maximum total alignment. Then we initialize topology structures to capture inter-modal correspondence between aligned feature nodes. The initialized structure is dynamically updated through topology reasoning for several steps to allow comprehensive and representative information propagation. Finally, we leverage a residual connection between learned input representations and reasoning scores based on cross-modal multi-head attention to realize a complementation.
Our contributions are summarized as follows:

\begin{itemize}
\item We notice the challenging problem of detecting implicit harmful memes and leverage optimal transport kernel method to model the cross-modal aligning task. To the best of our knowledge, this is the first work to incorporate such kernel methods with transportation theory in multimodal hate content detection.     

\item We implement topology-aware optimal transport to capture inter-modal correspondence, which establishes credible alignment based on transported feature nodes, and perform dynamic topology reasoning to allow comprehensive and representative information propagation. 
\item Without any external knowledge, our model achieves state-of-the-art performance by a large margin on two publicly available benchmarks. The subsequent visualization studies further confirm our model's superiority in capturing implicit cross-modal alignment.
\end{itemize}

\section{Related Work}
\subsection{Hate Content Detection}

With the prevalence of social media platforms, automated identification of hate content has received academic attention. Researchers from diverse communities have explored this challenging work \cite{DBLP:conf/coling/FortunaFPRN18}, and produced a large number of benchmark datasets \cite{DBLP:conf/hicss/BretschneiderP17,DBLP:journals/corr/RossRCCKW17,DBLP:journals/corr/abs-2108-05927}. Previous feature-engineering methods \cite{DBLP:journals/jetai/MalmasiZ18,DBLP:conf/sigdial/MehdadT16} mainly extracted and organized lower-level features, like $n$-gram and sentiment features. Nowadays, DNN-based methods have garnered better results by aggregating latent semantic features \cite{DBLP:conf/esws/ZhangRT18} or fine-tuning large pre-trained models \cite{DBLP:conf/acl/TekirougluCG20} like GPT-2. Although the existing methods of hateful content detection have yielded considerable experimental progress and commercial application, they just focused on text-based hateful content, neglecting the abundant multimedia UGC.

\subsection{Multimodal Hate Content Detection}
Frequent and repetitive exposure to multimodal harmful content will increase personal prejudice against external groups and further affect the status of socially disadvantaged groups. Enabling explorations were made for flourishing hateful meme detection studies by publishing benchmarks. For instance, Facebook had proposed the Hateful Memes Challenge \cite{DBLP:conf/nips/KielaFMGSRT20}, which encouraged researchers to identify the targeted harmful categories (e.g., race and sex). \cite{zia2021racist} proposed to classify memes beyond hateful and confirm the type of attack (e.g., contempt and slur). Recent MOMENTA \cite{DBLP:conf/emnlp/PramanickSDAN021} further extended the limited harmful categories and proposed two benchmarks related to COVID-19 and US politics.
Approaches of multimodal hateful memes classification \cite{DBLP:conf/nips/KielaFMGSRT20,suryawanshi2020multimodal} adopted early fusion techniques to incorporate linguistic and visual signals at input space and yield sub-optimal results.
Advanced systems fine-tuned large scale pre-trained unimodal and multimodal models such as VisualBERT \cite{DBLP:journals/corr/abs-1908-03557}, UNITER \cite{DBLP:journals/corr/abs-1909-11740}, VILLA \cite{gan2020large}, and ensemble thereof. More recently, DisMultiHate \cite{DBLP:conf/mm/LeeCFJC21} disentangled target entities to certain categories of hate (i.e., gender, race, etc.) in multimodal memes to improve classification and explainability. MOMENTA \cite{DBLP:conf/emnlp/PramanickSDAN021} systematically analyzed the local and the global perspective of the input meme and performed cross-modal attention in general level to detect harmful memes. However, aforementioned rivaling approaches relies on external demographic knowledge and have limited effectiveness in detecting harmful content beyond categories of gender or race, especially memes with implicit modal alignment. This paper aims to address this gap by proposing a general framework with optimal transported graph.

\begin{figure*}[!htbp]
    \centering
    \includegraphics[width=1\textwidth]{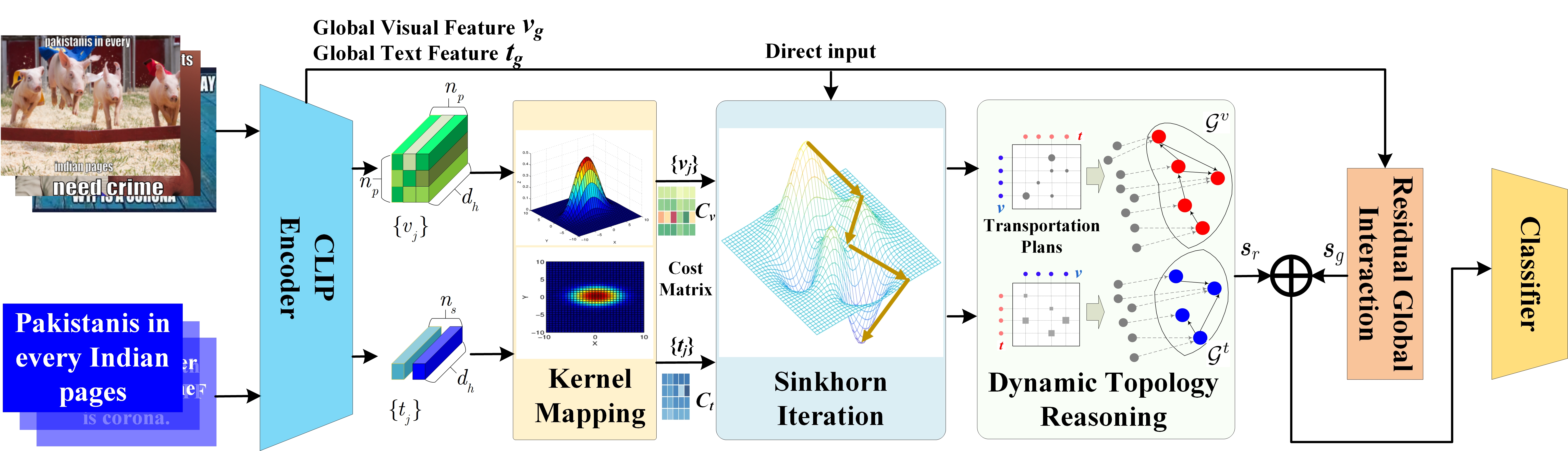}
    \caption{The model architecture of our TOT. Firstly, the representations of images and text can be learned through CLIP. Secondly, Kernel Mapping stage formulates the transportation cost matrix to optimize. Thirdly, we perform topology aligning and output two graphs with aligned representation. Finally, TOT conducts reasoning to calculate classification scores.}
    \label{tot}
\end{figure*}
\subsection{Optimal Transport }
Optimal transport is a well-studied topic in optimization theory, the original aim of which is to find the best transportation plan between two data distributions with minimum  cost. Recently, OT has indeed been widely used in alignment problems by researchers from various communities including computer vision and natural language processing \cite{DBLP:conf/aistats/GraveJB19}.  VOLT \cite{DBLP:conf/acl/XuZGZL20} formulated the quest of vocabularization as an optimal transport (OT) problem by finding the optimal transport matrix from the character distribution to the vocabulary token distribution. \cite{DBLP:conf/acl/ChenWTSCZWZC19} used the transport plan as an ad-hoc attention score in the context of network embedding to align data modalities. MuLOT \cite{DBLP:conf/wacv/PramanickRP22} utilized optimal transport based domain adaptation to learn strong cross-modal dependencies for sarcasm and humor detection. \cite{DBLP:conf/cvpr/GeLLYS21} innovatively revisited the label assignment from a global perspective and proposed to formulate the assigning procedure as an optimal transport (OT) problem. However, none of these studies have exploited optimal transport to learn element-level alignment representation as graph nodes. This paper contributes to memes analysis by proposing a novel framework TOT, which performs multimodal aligning for implicit harmful memes and constructs topology alignment graphs to allow dynamic topology reasoning.

\section{Methods}

\subsection{Input Representation Learning}

Unlike previous works that leveraged isolated encoders for visual and linguistic signals \cite{yu2021learning}, we choose CLIP \cite{radford2021learning} to learn the unified representations for input memes samples. CLIP is a visual-linguistic model pre-trained on 400M image–text pairs from the Internet through contrastive learning, which offers excellent zero-shot capabilities in capturing semantics for image-text inputs. Towards proposed TOT framework, we feed a meme’s image and its OCR-extracted text to  CLIP model initialized with pre-trained weights, then four embeddings $\boldsymbol{v}_{g}, \boldsymbol{t}_{g}, \boldsymbol{V}, \boldsymbol{T}$ from the multimodal encoder are extracted, where $\boldsymbol{v}_{g}, \boldsymbol{t}_{g} \in \mathbb {R}^{d_{h}}$ are global representations for multimodal input, and $\boldsymbol{V} \in \mathbb {R}^{n_{p}^{2} \times d_{h}}$, $\boldsymbol{T} \in \mathbb {R}^{n_{s} \times d_{h}}$ are sequence vectors containing local image and text information respectively. Note that $n_{p} \times n_{p}$ is the patch numbers per image and $n_{s}$ is the word numbers per sentence.

\subsection{Transportation Problem Formulating}
This section exploits the Gaussian kernel method \cite{mialon2021trainable} to formulate the cross-modal alignment problem as solutions for optimal transportation plans.

Based on the CLIP local visual feature sequence $\boldsymbol{V} = \{\boldsymbol{v}_{1},\boldsymbol{v}_{2},\dots,\boldsymbol{v}_{n_{p}^{2}}\}$  and local linguistic feature sequence $\boldsymbol{T} = \{\boldsymbol{t}_{1},\boldsymbol{t}_{2},\dots,\boldsymbol{t}_{n_{s}}\}$, we can acquire the cost matrix $\mathbf{C}$ towards transportation problem through a kernel method. Let $\kappa$ be the Gaussian kernel with reproducing kernel Hilbert space (RKHS) $\mathcal{H}$ and its associated kernel embedding $\varphi$ : $\mathbb{R}^{d} \rightarrow \mathcal{H}$. Then the $n_{s} \times n_{p}^{2} $ cost matrix $\mathbf{C}$ can be obtained by calculating the comparisons $\kappa(\boldsymbol{v}_{i},\boldsymbol{t}_{j})$ before detailed alignment.

With the cost matrix $\mathbf{C}$, the transportation plan from image to text, denoted by the $n_{s} \times n_{p}^{2} $ matrix ${\mathbf{P}(\boldsymbol{V,T})}$ is then defined as the unique solution of:

\begin{gather}
\mathop{min}\limits_{\mathbf{P}\in U}\sum\limits_{ij}{\mathbf{C}_{ij}\mathbf{P}_{ij}-\varepsilon H(\mathbf{P})}\\
H(\mathbf{P})=-\sum_{ij}{\mathbf{P}_{ij}(log(\mathbf{P}_{ij})-1)}
\end{gather}
where $\mathbf{C}$ represents the pairwise costs to align the elements of $\boldsymbol{V}$ and $\boldsymbol{T}$,  $H(\mathbf{P})$ refers to the entropy regularization with parameter $\epsilon$ to control the sparsity of $\mathbf{P}$, $U$ is the space of admissible couplings.


\subsection{Topology-Aware Optimal Transport}
This section describes the key module of proposed  TOT framework, which is designed to align multimodal representations based on transportation plans, and then establish topology structures to capture inter-modal correspondence as candidate edges among aligned feature nodes.

Based on transportation plan ${\mathbf{P}(\boldsymbol{V,T})}$, the aligned image nodes $\mathcal{G}^{v} = \{\boldsymbol{g}_{1}^{v},\boldsymbol{g}_{2}^{v},\dots \boldsymbol{g}_{n_{p}^{2}}^{v} \}$ can be acquired by:

\begin{gather}
\begin{aligned}
\mathcal{G}^{v} &= \sqrt{n_{p}^{2}}\times \left(\sum\limits_{i=1}^{n_{p}^{2}}{\mathbf{P}_{i1}\varphi(\boldsymbol{v}_{i}),\dots,\mathbf{P}_{ip}\varphi(\boldsymbol{v}_{n_{p}^{2}})}\right)\\
&= n_{p} \times \mathbf{P(V,T)}^{T}\varphi(\mathbf{V})
\end{aligned}
\label{equation_trans2}
\end{gather}
where the global visual representation $\boldsymbol{v}_{g}$ is appended as the first node of the aligned image nodes. To establish a topology structure for comprehensive alignment reasoning, we compute the candidate edge from $\boldsymbol{g}_{i}^{v} \in \mathcal{G}^{v}$ to $\boldsymbol{g}_{j}^{v} \in \mathcal{G}^{v}$ as:
\begin{equation}
 e(\boldsymbol{g}_{i}^{v},\boldsymbol{g}_{j}^{v}, \mathbf{W}_{Q}, \mathbf{W}_{K}) = \frac{exp((\mathbf{W}_{Q}\boldsymbol{g}_{i}^{v})(\mathbf{W}_{K}\boldsymbol{g}_{j}^{v}))}{\sum_{j}{exp((\mathbf{W}_{Q}\boldsymbol{g}_{i}^{v})(\mathbf{W}_{K}\boldsymbol{g}_{j}^{v}))}}
 \label{edges}
\end{equation}
where $\mathbf{W}_{Q} \in \mathbb{R}^{d_{h} \times h}$ and $\mathbf{W}_{K} \in \mathbb{R}^{d_{h} \times h}$ are linear transformations for generating query and key vector to compute candidate edges. Note that during edge calculation process, our topology image graph is directed and thus allows complex information propagation.

Similarly, the aligned text nodes
$\mathcal{G}^{t} = \{\boldsymbol{g}_{1}^{t},\boldsymbol{g}_{2}^{t},\dots \boldsymbol{g}_{n_{s}}^{t} \}$ can be acquired
based on the transportation plan $\mathbf{P(T,V)}$ from text to image:
\begin{gather}
\begin{aligned}
\mathcal{G}^{t} &= \sqrt{n_{s}}\times \left(\sum\limits_{i=1}^{n_{s}}{\mathbf{P}_{i1}\varphi(\boldsymbol{t}_{i}),\dots,\mathbf{P}_{ip}\varphi(\boldsymbol{t}_{n_{s}})}\right)\\
&= \sqrt{n_{s}} \times \mathbf{P(T,V)}^{T}\varphi(\mathbf{T})
\end{aligned}
\label{equation_trans}
\end{gather}
where the edge calculation of text nodes is based on equation \ref{edges}.
With the constructed TOT graphs, we perform dynamically topology reasoning to capture the inner-modal correspondence by iteratively updating  graph nodes and connected edges. The nodes $\boldsymbol{x}_{i}^{n}$ at $n$-th step   are updated as:
\begin{gather}
    \overline{\boldsymbol{x}}_{i}^{n+1} = \sum_{j}{e(\boldsymbol{x}_{i}^{n},\boldsymbol{x}_{j}^{n}, \mathbf{W}_{Q}^{n},\mathbf{W}_{K}^{n}) \cdot \boldsymbol{x}_{j}^{n}} \\
    \boldsymbol{x}_{i}^{n+1} = \textrm{ReLU}(\mathbf{W}_{a}^{n}  \overline{\boldsymbol{x}}_{i}^{n+1} + b_{a}^{n})
\end{gather}
where $\boldsymbol{x}_{i}^{n}$ are initialized by taking the $i$-th node from graph $\mathcal{G}_{v}$ or $\mathcal{G}_{t}$ at  $n$ = 0, and $\mathbf{W}_{Q}^{n},\mathbf{W}_{K}^{n},\mathbf{W}_{a}^{n},b_{a}^{n}$ are learnable parameters for $n$-th graph layer. After each step of topology reasoning, the  current node $\boldsymbol{x}_{i}^{n}$ is replaced with $\boldsymbol{x}_{i}^{n+1}$. 
When $N$ iterations stop, the global node is fed to a fully-connected layer to infer the reasoning score $\boldsymbol{s}_{r} \in \mathbb{R}^{d_{h}}$.

\subsection{Residual Global Interaction}

To complement the topology reasoning results and infer harmfulness from global perspective, we add residual connection between CLIP global representation and classification scores based on multi-head cross model attention.

Specifically, for the $i$-th head attention, the multimodal input is interacted based on the dot-product attention mechanism:
\begin{gather}
\textrm{ATT}_{i}(\boldsymbol{v}_{g},\boldsymbol{t}_{g})  =\sigma \left(\frac{[\mathbf{W}_{Q_{i}}\boldsymbol{v}_{g}]^{T}  [\mathbf{W}_{K_{i}}\boldsymbol{t}_{g}]}{\sqrt{d_{h}/m}}\right)\mathbf{W}_{V_{i}} \boldsymbol{t}_{g}
\label{equation_head}
\end{gather}
where $\{\mathbf{W}_{Q_{i}},\mathbf{W}_{K_{i}},\mathbf{W}_{V_{i}}\} \in \mathbb{R}^{d_{h}/m \times d_{h}}$ are learning parameters corresponding to queries, keys and values respectively.
Then the output of $m$ heads attention is concatenated together, followed by linear transformation and residual connection to get the multimodal representation $m_{r}$:
\begin{equation}
\boldsymbol{m}_{r} =\boldsymbol{t}_{g} + \mathbf{W}_m [\textrm{ATT}_{1}, \textrm{ATT}_{2}, ..., \textrm{ATT}_{m}]
\label{equation_multihead}
\end{equation}

With the global multimodal representation, a MLP layer is calculated to acquire the residual global score $\boldsymbol{s}_{g} \in \mathbb{R}^{d_{h}}$.

\subsection{Training Objectives}
With the residual global score $\boldsymbol{s}_{g}$ and topology reasoning score $\boldsymbol{s}_{r}$, we get the final multimodal hidden states $\boldsymbol{s}_f \in \mathbb{R}^{d_h}$ by weighted sum of both $\boldsymbol{s}_{g}$ and $\boldsymbol{s}_{r}$. Then the final hidden states are fed into a linear function followed by a softmax function for the classification:
\begin{equation}
    p(y\vert \boldsymbol{s}_{f}) = \text{softmax}(\mathbf{W_{c}}[(1-\gamma)\cdot s_{g} + \gamma \cdot s_{r}] + b_{c})
\end{equation}
where $\mathbf{W}_{c}\in \mathbb{R}^{c \times d_h}$, $c$ is the category number of dataset. 

The model parameters are optimized through back propagation with minimizing the standard cross-entropy loss function as the objective:
\begin{equation}
    \mathcal{L} = -\frac{1}{\lvert \mathcal{D}
    \rvert}\sum_{k=1}^{\lvert\mathcal{D}\rvert}\mathbf{log} p(y^{k}\vert \mathbf{s}_{m}^{k})
\label{loss_equation}
\end{equation}

\section{Experiment}
To verify the superiority of our proposed approach, in this section we firstly describe the evaluation settings. Then we display the model performance on two benchmarks of our TOT against other state-of-the-art unimodal and multimodal approaches. Finally, we conduct qualitative studies to analyze TOT's superiority in aligning clues from multiple modalities. We also display the error cases to discuss limitations of our model. 

\subsection{Datasets Settings}
\subsubsection{Datasets}
We evaluate our model on two publicly available harmful memes detection datasets, Harm-C and Harm-P, which consist of real-world memes relate to COVID-19 and US politics, respectively. The statistic information of the two benchmarks are displayed in Table \ref{dataset}. To make a fair comparison with the previous work, we report three metrics: Accuracy (Acc), Macro-F1 (F1), and Macro-Averaged Mean Absolute Error (MMAE).

\begin{table}[!htbp]
	\centering
    \normalsize
		\begin{tabular}{cccccc}
			\toprule
			
	Dataset  &Harmfulness	& Train  & Test & Valid & Total \\
    \midrule
		\multirow{4}{*}{Harm-C}	&Very  & 182  & 21 & 10  & 213  \\
			
			&Partially & 882  & 103  & 51  & 1036  \\
			&Non & 1949  & 230 & 116  & 2295   \\	
			&Total & 3013  & 354  & 177 & 3544 \\
	\midrule
    		\multirow{4}{*}{Harm-P} &Very & 216  & 25 & 17  & 258  \\
    		
    		&Partially & 1270  & 148  & 69  & 1487  \\
    		&Non & 1534  & 182 & 91  & 1807   \\	
    		&Total & 3020  & 355  & 177 & 3552 \\
			\bottomrule
		
	\end{tabular}
	\caption{Statistics of Harm-C and Harm-P dataset.}
	\label{dataset}
\end{table}	

\begin{table*}[!ht]
    \large
	\centering

	\setlength{\tabcolsep}{0.9mm}{
    \renewcommand\arraystretch{1.3}
	\normalsize
		\begin{tabular}{clcccccccccccc}
			\toprule 
			\multirow{3}{*}{Modality}
			&\multirow{3}{*}{Model}  &\multicolumn{6}{c}{2-class Harmful
			Meme Detection}
			&\multicolumn{6}{c}{3-class Harmful
			Meme Detection}
			\\
			\cmidrule{3-8}    
			\cmidrule{9-14}
		    && \multicolumn{3}{c}{Harm-C} 
		    & \multicolumn{3}{c}{Harm-P} 
		    & \multicolumn{3}{c}{Harm-C} 
		    & \multicolumn{3}{c}{Harm-P} 
		    \\

			&& Acc    & F1    &MMAE  & Acc    & F1  &MMAE & Acc    & F1  &MMAE & Acc    & F1  &MMAE  \\
			\midrule

    		\multirow{3}{*}{Text} &BERT  &70.17 & 66.25 &0.2911 & 80.12 & 78.35 &0.1660 &68.93 & 48.72 &0.5591 &74.55 & 54.08 &0.7742
			\\
			 &RoBERTa  &70.77 & 66.40 &0.2923 & 80.74 & 78.64 &0.1644 &70.11 & 48.94 &0.5584 &74.87 & 54.33 &0.7771
			\\
			&Bertweet  &71.32 & 67.30 &0.2842 & 81.88 & 79.34 &0.1573 &71.65 & 49.82 &0.5512 &75.22 & 54.87 &0.7652
			\\
			\cmidrule{1-14}

		\multirow{2}{*}{Image}	&DenseNet &68.42 & 62.54 &0.3125 & 74.05 & 73.68 &0.1845 &65.21 & 42.15 &0.6326 &71.80 & 50.98 &0.8388
			\\
		
			&ResNet &68.74 & 62.97 &0.3114 & 73.14 & 72.77 &0.1800 &65.29 & 43.02 &0.6264 &71.02 & 50.64 &0.8900
			\\

		    \cmidrule{1-14}

		
			\multirow{5}{*}{Multimodal}&MMBT  &73.48 & 67.12 &0.3258 & 82.54 & 80.23 &0.1413 &68.08 & 50.88 &0.6474 &78.14& 58.03 &0.7008
			\\
			&ViLBERT &78.53  & 78.06 &0.1881  & 87.25 & 86.03 &\textbf{0.1276}  &75.71 & 48.82 &0.5329 &84.66 & 64.70 &0.6982
			\\
			&Visual BERT  &81.36 & 80.13 &0.1857 & 86.80  & 86.07 &0.1318 &74.01 & 53.85 &0.5303 &84.02 & 63.68 &0.7020
		    \\
			&CLIP  &78.10 & 77.64 &0.2010 & 84.02 & 83.85 & 0.1508 &71.05 & 45.55 &0.5887 &80.75 &60.23 &0.7058
			\\

    			&MOMENTA  &\textbf{83.82} & \textbf{82.80} &\textbf{0.1743} & \textbf{89.84} & \textbf{88.26} &0.1314 &\textbf{77.10} & \textbf{54.74} &\textbf{0.5132} &\textbf{87.14} & \textbf{66.66} &\textbf{0.6805}
			\\
	        \cmidrule{1-14}
	        \multirow{4}{*}{Ours}
			&TOH  &82.16 & 81.36 &0.1877 & 88.56 & 86.93 &0.1428 &75.44 & 53.08 &0.5276 &84.71 & 64.21 &0.6957
			\\
			&TOA  &83.55 & 82.14 &0.1813 & 89.17 & 87.44 &0.1367 &76.25 & 53.74 &0.5233 &85.34 & 65.39 &0.6864
			\\
			&COT  &86.48 & 85.28 &0.1697 & 90.24 & 90.88 &0.1278 &81.56 & 54.21 &0.5147 &86.37 & 69.04 &0.6742
			\\
			&TOT  &\textbf{87.01} & \textbf{85.93} &\textbf{0.1634} & \textbf{91.55} & \textbf{91.29} &\textbf{0.1245} &\textbf{82.76} & \textbf{55.38} &\textbf{0.5027} &\textbf{88.61} & \textbf{71.54} &\textbf{0.6697}
			\\
			\cmidrule{1-14}
			&$\triangle_{\text{TOT-best\_model}}$  &3.19 & 3.13 &0.0109 & 1.71 & 3.02 &0.0031 &5.66 & 0.64 &0.0105 &1.47 & 4.88 &0.0108
			\\
			\bottomrule
			
	\end{tabular}}
	\caption{Overall performance of proposed TOT model against a set of baselines. For Acc ($\uparrow$) and F1 ($\uparrow$), the larger values are better , while for MMAE($\downarrow$), smaller values are better.}
	\label{main_ex}
\end{table*}

\subsubsection{Baselines}
For both Harm-C and Harm-P, we compare the model performance with a set of strong baselines:
\begin{itemize}
    \item \textbf{BERT, RoBERTa, Bertweet:} For the language models, we choose influential pre-trained models: BERT \cite{kenton2019bert}, RoBERTa \cite{DBLP:journals/corr/abs-1907-11692} and Bertweet \cite{nguyen2020bertweet} .

    \item \textbf{DenseNet, ResNet:} For the visual models, we choose two famous models pre-trained on ImageNet: DenseNet \cite{huang2017densely} and ResNet \cite{he2016deep}.
    \item \textbf{MMBT:} This is a multimodal bitransformer \cite{DBLP:conf/nips/KielaBFT19}, which is able to capture the intra-modal and the inter-modal dynamics of the two modalities.
    \item \textbf{ViLBERT:} This is a robust model trained on an intermediate multimodal objective (conceptual captions) \cite{sharma2018conceptual} with task-agnostic representations.
    \item \textbf{Visual BERT:} This is a visual BERT \cite{DBLP:journals/corr/abs-1908-03557} pre-trained on the COCO dataset \cite{DBLP:conf/eccv/LinMBHPRDZ14}.
    \item \textbf{MOMENTA:} This is a novel multimodal deep neural network using background context to detect hate.
\end{itemize}

\subsubsection{Implementation Details}
For each meme, we take the $d_{h} = 512$ to extract representations.  We limit the max lengths of image and text feature sequence by setting $n_{p}^{2} = 49$ and $n_{s} = 77$ respectively, which adopts the same configuration as the pre-training process. For the Kernel Mapping, we take Gaussian kernel with $\epsilon = 0.1$ and set the max numbers of sinkhorn iteration for as 3. For the  topology reasoning, we take a configuration of reason step $N=3$ and a dimension $h = 256$ for node vectors.


\subsection{Quatitative Analysis}	

In this section, we report the 2-class and 3-class detection results on Harm-C and Harm-P datasets to demonstrate the superiority of proposed approach from multiple perspectives. Note that we implement three variations (TOH, TOA, COT) from TOT model to demonstrate superiority. TOH is a variation by replacing the topology aligning module with heuristic aligning \cite{diao2021similarity}, and TOA replaces the topology aligning module with attentional aligning. COT directly concatenates the multimodal features.

\subsubsection{2-class Detection Results}
The left part of Table \ref{main_ex} reports the experimental results on Harm-C and Harm-P as the binary case.
Firstly, it can be observed Bertweet achieves relative outstanding performance in the unimodal approaches. 
A possible reason for the phenomenon could be due to the excellent social media capability of Bertweet, which is pre-trained on Twitter corpus.  
Secondly, the multimodal approaches achieve overall superior performance against unimodel ones.
And approaches which finetunes the large-scale pre-trained visual-linguistic models are better compared Intermidiate Fusion, as the pre-trained models are able to capture stronger inter-modal dynamics of the two modalities. 
Thirdly, we can see that our proposed TOT model outperforms the existing methods by a larger margin, with  3.13\% absolute points of M-F1 improvement on Harm-C and 3.02\% points on Harm-P. The other two metrics of Acc and MMAE also get improved by a competitive margin, demonstrating the effectiveness of our method.

\subsubsection{3-class Detection Results}
To further validate models' ability in fine-grained harmful contents detection, we also report the 3-class classification results in Table \ref{main_ex}.
Similar phenomena can be observed compared to the binary case. Our proposed TOT outperforms other methods by a larger margin, with 5.66\% absolute Accuracy points on Harm-C and 4.88\% M-F1 points on Harm-P. It is worth noted that competitive performance can be also achieved by COT, which directly concatenates the alignment representations instead of building transported graphs, further validating
the superiority of our models. 

\subsection{Ablation Studies}
In this section, we have conducted extensive ablation studies to verify the effectiveness of the three modules implemented in our method, including optimal transported image/text graphs (OTI/OTT), dynamic topology reasoning (DTOR) and Residual Global Interaction (REG). All the comparative experiments are conducted in the binary classification task.

\subsubsection{Optimal Transported Graphs Ablation}

\begin{table}[!htbp]
\centering

\normalsize
\begin{tabular}{cccccc}

\toprule
Datasets & OTT & OTI & DTOR & Acc &M-F1 \\
\hline

\multirow{6}{*}{Harm-C}   
    & × & &  &86.41 &85.33          
\\ 
    &  & × & &86.17 &85.07          
\\ 
    & & &×  &85.78 &84.46         
\\ 
    &× & &× &84.93 &83.76            
\\ 
    & &× &× &84.82 &84.64            
\\ 
    &× &× &× &83.74 &82.46
\\ 
\hline
\multirow{6}{*}{Harm-P}
    & × & & &90.34 & 90.17
\\
    &  & × & &90.07 & 89.88             
\\ 
    & & &× &89.71 & 89.43            
\\ 
    &× & &× &87.46 & 87.27             
\\ 
    & &× &× &87.39 & 87.14               
\\ 
    &× &× &× &86.73 & 86.24
\\
\bottomrule

\end{tabular}
\caption{Ablation results of transported graphs.}
\label{OTG_ablation}
\end{table}

Table \ref{OTG_ablation} shows the binary classification results of different settings for optimal transported graphs and topology graph reasoning module (DTOR). For the models without DTOR, we adopt a MLP layer instead to output the topology reasoning scores. Based on the results we can make observations that the two transported graphs OTT and OTI are both helpful for classification. OTT and OTI contribute to the Accuracy score at 1.08\% and 1.09\% absolute points in Harm-C. And when the two graphs are exploited together, they make further contributions by performing additional complementary effect. And the implementation of DTOR brings performance improvement at 1.2\% average absolute points in accuracy based on OT graphs.

\subsubsection{Residual Global Interaction Ablation}
This subsection explores the complementary effect of classification scores by ablating topology reasoning score and residual global score. We also report different training strategies for these two modules: joint learning and independent learning, which aims to further verify their complementary effect during learning. The ablation results are shown in Table \ref{reasoning_ablation}. We can observe that the scores from these two modules together achieve better performance compared to any single score (1.57\% points in Acc and 1.45\% points in M-F1 for Harm-C dataset), which well confirm the effectiveness of REG module. Besides, the joint training process contributes more compared to separately training by an improvement of 0.59\% in Acc, as the two modules perform better complementary effect when training jointly.

\begin{table}[!htbp]
\centering

\footnotesize
\setlength{\tabcolsep}{\tabcolsep}{
\begin{tabular}{ccccccc}

\toprule
Datasets  & DTOR & REG &Split &Joint &Acc &M-F1 
\\ 
\hline
\multirow{4}{*}{Harm-C}   
     &\checkmark & &\checkmark & 
     & 85.44 &84.48
\\ 
     & &\checkmark &\checkmark &  
     & 84.37 &83.27
\\ 
     &\checkmark &\checkmark &\checkmark &  & 86.42 &85.67      
\\ 
     &\checkmark &\checkmark & &\checkmark    & 87.01 &85.93
\\ 
\hline
\multirow{4}{*}{Harm-P}   
     &\checkmark & &\checkmark & 
     &89.83 &89.44
\\ 
     & &\checkmark &\checkmark &        &87.27 &87.03
\\ 
     &\checkmark &\checkmark &\checkmark &  &91.15 &90.87       
\\ 
     &\checkmark &\checkmark & &\checkmark  &91.55 &91.29
\\
\bottomrule
\end{tabular}
}
\caption{Abaltion results of reasoning modules.}
\label{reasoning_ablation}
\end{table}

\subsection{Qualitative Results}
The ability to formulate and capture cross-modal alignment for deciphering implicit hate is a core contribution in our TOT model. Toward this research goal, we design qualitative analysis including case study and visualizations.

\begin{figure*}[!htbp]
    \centering
    \includegraphics[width=0.70\textwidth]{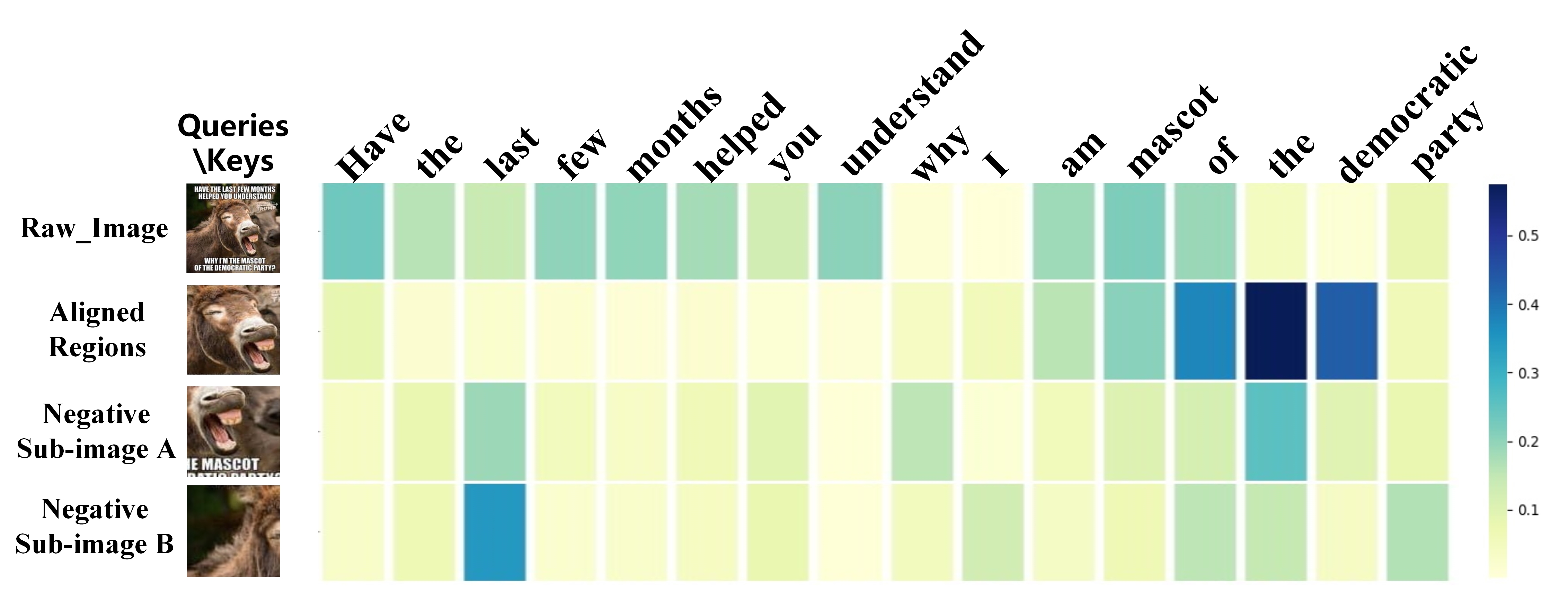}
    \caption{The visualizations of edge values in TOT. Raw\_Image denotes the input meme image, Aligned Regions denotes the manually selected image regions containing alignments with text information for disseminating harm. The two negative samples are randomly selected sub-image incorporated for comparison.}
    \label{heat}
\end{figure*}

\begin{figure*}[!ht]
    \centering
    \includegraphics[width=0.83\textwidth]{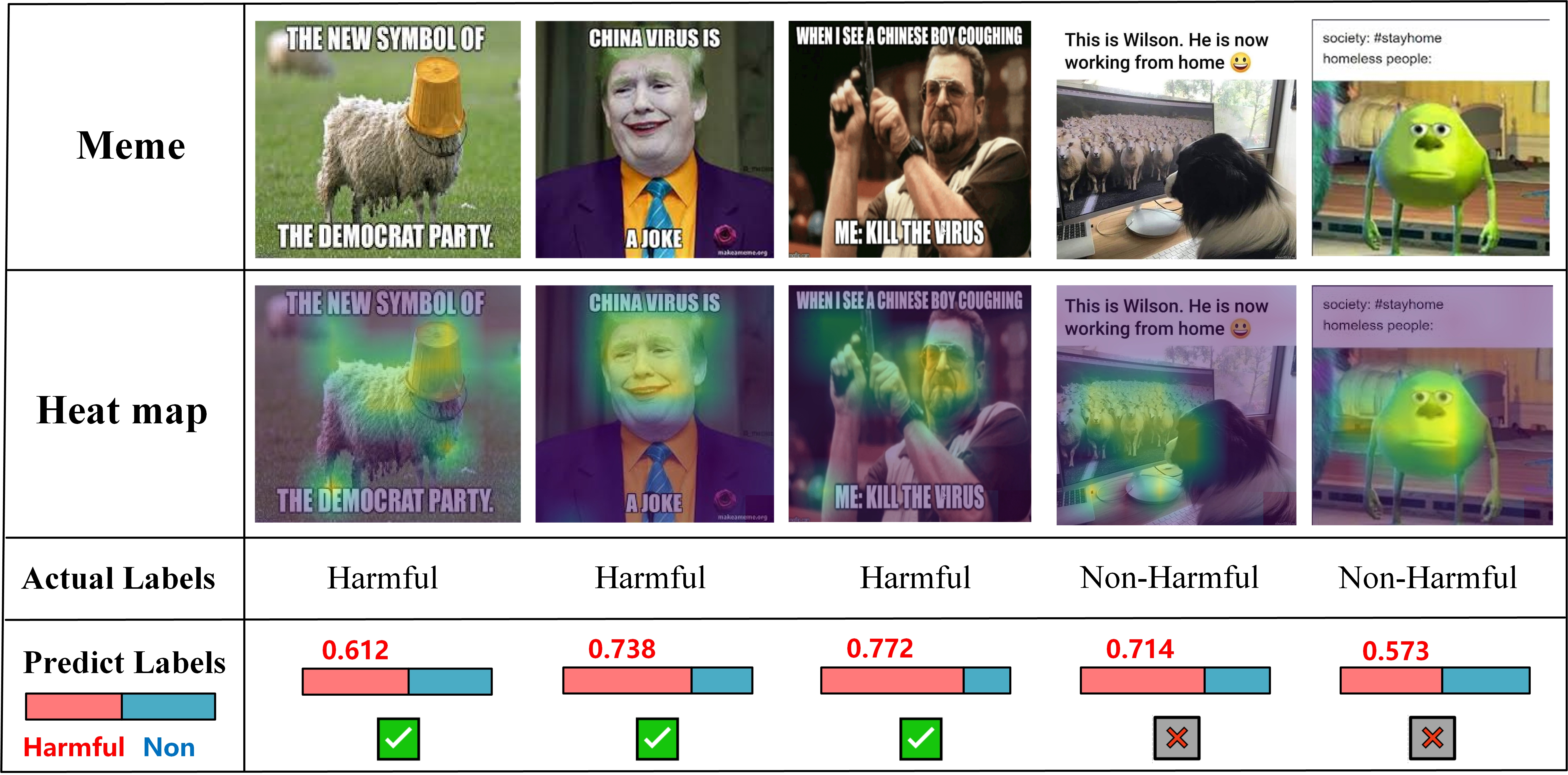}
    \caption{Case studies of memes from Harm-C and Harm-P datasets.}
    \label{case}
\end{figure*}
\subsubsection{Visualizations}
We visualize the edge values after $N$ step iterations of topology reasoning between global text nodes and other candidate nodes. Figure \ref{heat} displays what information the ToT module will aggregate given different visual features as input queries. The representations of visual regions are acquired by taking the average values of all the image patches that contains the target. For the offensive memes,  TOT module captures implicit alignment between the meaningful instances ('mascot of the democratic party') and the slovenly donkey in the picture. Moreover, negative samples are exploited as queries to make comparison. It is observed that when exploiting negative samples, the final global node randomly aggregates meaningless information, as the misaligned representations by transportation plans influence the performance of topology reasoning. 


\subsubsection{Case Study}
As shown in Figure \ref{case}, we display several memes from the leveraged datasets, together with the heat map of edge values between global visual nodes and other candidate nodes.  The text information in first meme is an innocent description, while its picture shows offensive content, which satirizes 'The Democrat Party' in the sentence. The second meme disseminates extreme multimodal harm reflected by the 'CHINA VIRUS' in the text and the word 'JOKE' with double meaning, which is implicitly aligned with the vicious painted picture of Trump. The third picture has a discrimination harm. The two sentences seem to describe irrelevant things, but when they are put together with the image, in which a man is firing a gun, the alignment between 'KILL THE VIRUS' and 'A CHINESE BOY COUGHING' is established, and results in the extreme harm on race discrimination. These three cases demonstrate the necessity of aligning multimodal features.

Besides, we are also interested in exploring the limitations of our TOT. The wrongly classified memes are displayed in Figure \ref{case}. The fourth example is a funny meme, which shows a cute sheepdog watching the sheep in the screen to do his herding. Note that Wilson is the name of the dog. However, the anthropomorphized expression in the sentence, together with the dog pictures misleads our TOT to make wrong predictions, which is quite similar to the dehumanizing hateful content. The similar situation appears in the fifth meme, which may indicate that well cross-modal alignment is necessary but not adequate for detecting harm. 


\section{Conclusion}
In this paper, we put forward a novel research problem of detecting implicit-aligned harmful memes, where this is the first work to incorporate such kernel methods with transportation theory in multimodal hate content detection. Specifically,  topology-aware optimal transport (TOT)  is conducted to formulate the cross-modal aligning problem as solutions for optimal transportation plans. With the dynamic topology reasoning module (DTOR), our TOT allows comprehensive information propagation based on the aligned information to make accurate reasoning. TOT is evaluated on two publicly available datasets, and our extensive experiments have shown that TOT outperforms
the state-of-the-art baselines. We have also conducted visualizations and case studies to empirically demonstrated TOT’s ability to capture implicit cross-modal alignment. While the error case shows that well cross-modal alignment is necessary but not adequate for detecting harm. We will incorporate a more advanced reasoning module for future improvements.

\clearpage
\section*{Acknowledgements}

This research was funded by the National Natural Science Foundation of China (62206267), and the Strategic Priority Research Program of the Chinese Academy of Sciences grant (Y835120378).

\bibliography{ref.bib}

\begin{thebibliography}{42}
\providecommand{\natexlab}[1]{#1}

\bibitem[{Bretschneider and Peters(2017)}]{DBLP:conf/hicss/BretschneiderP17}
Bretschneider, U.; and Peters, R. 2017.
\newblock Detecting Offensive Statements towards Foreigners in Social Media.
\newblock In \emph{HICSS}, 1--10.

\bibitem[{Chen et~al.(2020)Chen, Gan, Cheng, Li, Carin, and Liu}]{ChenG0LC020}
Chen, L.; Gan, Z.; Cheng, Y.; Li, L.; Carin, L.; and Liu, J. 2020.
\newblock Graph Optimal Transport for Cross-Domain Alignment.
\newblock In \emph{ICML}, volume 119, 1542--1553.

\bibitem[{Chen et~al.(2019{\natexlab{a}})Chen, Wang, Tao, Shen, Cheng, Zhang,
  Wang, Zhang, and Carin}]{DBLP:conf/acl/ChenWTSCZWZC19}
Chen, L.; Wang, G.; Tao, C.; Shen, D.; Cheng, P.; Zhang, X.; Wang, W.; Zhang,
  Y.; and Carin, L. 2019{\natexlab{a}}.
\newblock Improving Textual Network Embedding with Global Attention via Optimal
  Transport.
\newblock In \emph{ACL}, 5193--5202.

\bibitem[{Chen et~al.(2019{\natexlab{b}})Chen, Li, Yu, Kholy, Ahmed, Gan,
  Cheng, and Liu}]{DBLP:journals/corr/abs-1909-11740}
Chen, Y.; Li, L.; Yu, L.; Kholy, A.~E.; Ahmed, F.; Gan, Z.; Cheng, Y.; and Liu,
  J. 2019{\natexlab{b}}.
\newblock {UNITER:} Learning UNiversal Image-TExt Representations.
\newblock \emph{CoRR}, abs/1909.11740.

\bibitem[{Das, Wahi, and Li(2020)}]{das2020detecting}
Das, A.; Wahi, J.~S.; and Li, S. 2020.
\newblock Detecting hate speech in multi-modal memes.
\newblock \emph{arXiv preprint arXiv:2012.14891}.

\bibitem[{Diao et~al.(2021)Diao, Zhang, Ma, and Lu}]{diao2021similarity}
Diao, H.; Zhang, Y.; Ma, L.; and Lu, H. 2021.
\newblock Similarity reasoning and filtration for image-text matching.
\newblock In \emph{AAAI}, 1218--1226.

\bibitem[{Duan et~al.(2022)Duan, Chen, Tran, Yang, Xu, Zeng, and
  Chilimbi}]{DuanCTYXZC22}
Duan, J.; Chen, L.; Tran, S.; Yang, J.; Xu, Y.; Zeng, B.; and Chilimbi, T.
  2022.
\newblock Multi-modal Alignment using Representation Codebook.
\newblock In \emph{CVPR}, 15630--15639. IEEE.

\bibitem[{Fortuna et~al.(2018)Fortuna, Ferreira, Pires, Routar, and
  Nunes}]{DBLP:conf/coling/FortunaFPRN18}
Fortuna, P.; Ferreira, J.; Pires, L.; Routar, G.; and Nunes, S. 2018.
\newblock Merging Datasets for Aggressive Text Identification.
\newblock In \emph{COLING}, 128--139.

\bibitem[{Gan et~al.(2020)Gan, Chen, Li, Zhu, Cheng, and Liu}]{gan2020large}
Gan, Z.; Chen, Y.-C.; Li, L.; Zhu, C.; Cheng, Y.; and Liu, J. 2020.
\newblock Large-scale adversarial training for vision-and-language
  representation learning.
\newblock \emph{Advances in Neural Information Processing Systems}, 33:
  6616--6628.

\bibitem[{Ge et~al.(2021)Ge, Liu, Li, Yoshie, and
  Sun}]{DBLP:conf/cvpr/GeLLYS21}
Ge, Z.; Liu, S.; Li, Z.; Yoshie, O.; and Sun, J. 2021.
\newblock {OTA:} Optimal Transport Assignment for Object Detection.
\newblock In \emph{CVPR}, 303--312.

\bibitem[{Grave, Joulin, and Berthet(2019)}]{DBLP:conf/aistats/GraveJB19}
Grave, E.; Joulin, A.; and Berthet, Q. 2019.
\newblock Unsupervised Alignment of Embeddings with Wasserstein Procrustes.
\newblock In \emph{AISTATS}, volume~89, 1880--1890.

\bibitem[{He et~al.(2016)He, Zhang, Ren, and Sun}]{he2016deep}
He, K.; Zhang, X.; Ren, S.; and Sun, J. 2016.
\newblock Deep residual learning for image recognition.
\newblock In \emph{CVPR}, 770--778.

\bibitem[{Huang et~al.(2017)Huang, Liu, Van Der~Maaten, and
  Weinberger}]{huang2017densely}
Huang, G.; Liu, Z.; Van Der~Maaten, L.; and Weinberger, K.~Q. 2017.
\newblock Densely connected convolutional networks.
\newblock In \emph{CVPR}, 4700--4708.

\bibitem[{Kenton and Toutanova(2019)}]{kenton2019bert}
Kenton, J. D. M.-W.~C.; and Toutanova, L.~K. 2019.
\newblock BERT: Pre-training of Deep Bidirectional Transformers for Language
  Understanding.
\newblock In \emph{Proceedings of NAACL-HLT}, 4171--4186.

\bibitem[{Kiela et~al.(2019)Kiela, Bhooshan, Firooz, and
  Testuggine}]{DBLP:conf/nips/KielaBFT19}
Kiela, D.; Bhooshan, S.; Firooz, H.; and Testuggine, D. 2019.
\newblock Supervised Multimodal Bitransformers for Classifying Images and Text.
\newblock In \emph{NeurIPS}.

\bibitem[{Kiela et~al.(2020)Kiela, Firooz, Mohan, Goswami, Singh, Ringshia, and
  Testuggine}]{DBLP:conf/nips/KielaFMGSRT20}
Kiela, D.; Firooz, H.; Mohan, A.; Goswami, V.; Singh, A.; Ringshia, P.; and
  Testuggine, D. 2020.
\newblock The Hateful Memes Challenge: Detecting Hate Speech in Multimodal
  Memes.
\newblock In \emph{NeurIPS}, 2611--2624.

\bibitem[{Lee et~al.(2021)Lee, Cao, Fan, Jiang, and
  Chong}]{DBLP:conf/mm/LeeCFJC21}
Lee, R.~K.; Cao, R.; Fan, Z.; Jiang, J.; and Chong, W. 2021.
\newblock Disentangling Hate in Online Memes.
\newblock In \emph{MM}, 5138--5147.

\bibitem[{Li et~al.(2019)Li, Yatskar, Yin, Hsieh, and
  Chang}]{DBLP:journals/corr/abs-1908-03557}
Li, L.~H.; Yatskar, M.; Yin, D.; Hsieh, C.; and Chang, K. 2019.
\newblock VisualBERT: {A} Simple and Performant Baseline for Vision and
  Language.
\newblock \emph{CoRR}, abs/1908.03557.

\bibitem[{Lin et~al.(2014)Lin, Maire, Belongie, Hays, Perona, Ramanan,
  Doll{\'{a}}r, and Zitnick}]{DBLP:conf/eccv/LinMBHPRDZ14}
Lin, T.; Maire, M.; Belongie, S.~J.; Hays, J.; Perona, P.; Ramanan, D.;
  Doll{\'{a}}r, P.; and Zitnick, C.~L. 2014.
\newblock Microsoft {COCO:} Common Objects in Context.
\newblock In \emph{ECCV}, volume 8693, 740--755.

\bibitem[{Lippe et~al.(2020)Lippe, Holla, Chandra, Rajamanickam, Antoniou,
  Shutova, and Yannakoudakis}]{lippe2020multimodal}
Lippe, P.; Holla, N.; Chandra, S.; Rajamanickam, S.; Antoniou, G.; Shutova, E.;
  and Yannakoudakis, H. 2020.
\newblock A multimodal framework for the detection of hateful memes.
\newblock \emph{arXiv preprint arXiv:2012.12871}.

\bibitem[{Liu et~al.(2019)Liu, Ott, Goyal, Du, Joshi, Chen, Levy, Lewis,
  Zettlemoyer, and Stoyanov}]{DBLP:journals/corr/abs-1907-11692}
Liu, Y.; Ott, M.; Goyal, N.; Du, J.; Joshi, M.; Chen, D.; Levy, O.; Lewis, M.;
  Zettlemoyer, L.; and Stoyanov, V. 2019.
\newblock RoBERTa: {A} Robustly Optimized {BERT} Pretraining Approach.
\newblock \emph{CoRR}, abs/1907.11692.

\bibitem[{Malmasi and Zampieri(2018)}]{DBLP:journals/jetai/MalmasiZ18}
Malmasi, S.; and Zampieri, M. 2018.
\newblock Challenges in discriminating profanity from hate speech.
\newblock \emph{J. Exp. Theor. Artif. Intell.}, 30(2): 187--202.

\bibitem[{Mandl et~al.(2021)Mandl, Modha, Shahi, Jaiswal, Nandini, Patel,
  Majumder, and Sch{\"{a}}fer}]{DBLP:journals/corr/abs-2108-05927}
Mandl, T.; Modha, S.; Shahi, G.~K.; Jaiswal, A.~K.; Nandini, D.; Patel, D.;
  Majumder, P.; and Sch{\"{a}}fer, J. 2021.
\newblock Overview of the {HASOC} track at {FIRE} 2020: Hate Speech and
  Offensive Content Identification in Indo-European Languages.
\newblock \emph{CoRR}, abs/2108.05927.

\bibitem[{Maretic et~al.(2022)Maretic, Gheche, Chierchia, and
  Frossard}]{MareticGCF22}
Maretic, H.~P.; Gheche, M.~E.; Chierchia, G.; and Frossard, P. 2022.
\newblock fGOT: Graph Distances Based on Filters and Optimal Transport.
\newblock In \emph{AAAI}, 7710--7718.

\bibitem[{Mehdad and Tetreault(2016)}]{DBLP:conf/sigdial/MehdadT16}
Mehdad, Y.; and Tetreault, J.~R. 2016.
\newblock Do Characters Abuse More Than Words?
\newblock In \emph{SIGDIAL}, 299--303.

\bibitem[{Mialon et~al.(2021)Mialon, Chen, d'Aspremont, and
  Mairal}]{mialon2021trainable}
Mialon, G.; Chen, D.; d'Aspremont, A.; and Mairal, J. 2021.
\newblock A Trainable Optimal Transport Embedding for Feature Aggregation and
  its Relationship to Attention.
\newblock In \emph{ICLR}.

\bibitem[{Nguyen, Vu, and Nguyen(2020)}]{nguyen2020bertweet}
Nguyen, D.~Q.; Vu, T.; and Nguyen, A.-T. 2020.
\newblock BERTweet: A pre-trained language model for English Tweets.
\newblock In \emph{EMNLP}, 9--14.

\bibitem[{Pan et~al.(2020)Pan, Lin, Fu, Qi, and Wang}]{PanL0Q020}
Pan, H.; Lin, Z.; Fu, P.; Qi, Y.; and Wang, W. 2020.
\newblock Modeling Intra and Inter-modality Incongruity for Multi-Modal Sarcasm
  Detection.
\newblock In \emph{Findings of EMNLP}, 1383--1392.

\bibitem[{Pramanick, Roy, and Patel(2022)}]{DBLP:conf/wacv/PramanickRP22}
Pramanick, S.; Roy, A.; and Patel, V.~M. 2022.
\newblock Multimodal Learning using Optimal Transport for Sarcasm and Humor
  Detection.
\newblock In \emph{WACV}, 546--556.

\bibitem[{Pramanick et~al.(2021)Pramanick, Sharma, Dimitrov, Akhtar, Nakov, and
  Chakraborty}]{DBLP:conf/emnlp/PramanickSDAN021}
Pramanick, S.; Sharma, S.; Dimitrov, D.; Akhtar, M.~S.; Nakov, P.; and
  Chakraborty, T. 2021.
\newblock {MOMENTA:} {A} Multimodal Framework for Detecting Harmful Memes and
  Their Targets.
\newblock In \emph{EMNLP}, 4439--4455.

\bibitem[{Radford et~al.(2021)Radford, Kim, Hallacy, Ramesh, Goh, Agarwal,
  Sastry, Askell, Mishkin, Clark et~al.}]{radford2021learning}
Radford, A.; Kim, J.~W.; Hallacy, C.; Ramesh, A.; Goh, G.; Agarwal, S.; Sastry,
  G.; Askell, A.; Mishkin, P.; Clark, J.; et~al. 2021.
\newblock Learning transferable visual models from natural language
  supervision.
\newblock In \emph{ICML}, 8748--8763.

\bibitem[{Ross et~al.(2017)Ross, Rist, Carbonell, Cabrera, Kurowsky, and
  Wojatzki}]{DBLP:journals/corr/RossRCCKW17}
Ross, B.; Rist, M.; Carbonell, G.; Cabrera, B.; Kurowsky, N.; and Wojatzki, M.
  2017.
\newblock Measuring the Reliability of Hate Speech Annotations: The Case of the
  European Refugee Crisis.
\newblock \emph{CoRR}, abs/1701.08118.

\bibitem[{Sharma et~al.(2018)Sharma, Ding, Goodman, and
  Soricut}]{sharma2018conceptual}
Sharma, P.; Ding, N.; Goodman, S.; and Soricut, R. 2018.
\newblock Conceptual captions: A cleaned, hypernymed, image alt-text dataset
  for automatic image captioning.
\newblock In \emph{ACL}, 2556--2565.

\bibitem[{Suryawanshi et~al.(2020)Suryawanshi, Chakravarthi, Arcan, and
  Buitelaar}]{suryawanshi2020multimodal}
Suryawanshi, S.; Chakravarthi, B.~R.; Arcan, M.; and Buitelaar, P. 2020.
\newblock Multimodal meme dataset (MultiOFF) for identifying offensive content
  in image and text.
\newblock In \emph{TRAC@LREC}, 32--41.

\bibitem[{Tekiroglu, Chung, and Guerini(2020)}]{DBLP:conf/acl/TekirougluCG20}
Tekiroglu, S.~S.; Chung, Y.; and Guerini, M. 2020.
\newblock Generating Counter Narratives against Online Hate Speech: Data and
  Strategies.
\newblock In \emph{ACL}, 1177--1190.

\bibitem[{Velioglu, Rose, and Analytics(2020)}]{velioglu2020detecting}
Velioglu, R.; Rose, J.; and Analytics, S.~D. 2020.
\newblock Detecting Hate Speech in Memes Using Multimodal Deep Learning
  Approaches: Prize-winning solution to Hateful Memes Challenge.
\newblock \emph{arXiv preprint arXiv:2012.12975}.

\bibitem[{Xu et~al.(2021)Xu, Zhou, Gan, Zheng, and Li}]{DBLP:conf/acl/XuZGZL20}
Xu, J.; Zhou, H.; Gan, C.; Zheng, Z.; and Li, L. 2021.
\newblock Vocabulary Learning via Optimal Transport for Neural Machine
  Translation.
\newblock In \emph{ACL}, 7361--7373.

\bibitem[{Yu et~al.(2021)Yu, Xu, Yuan, and Wu}]{yu2021learning}
Yu, W.; Xu, H.; Yuan, Z.; and Wu, J. 2021.
\newblock Learning modality-specific representations with self-supervised
  multi-task learning for multimodal sentiment analysis.
\newblock In \emph{AAAI}, 10790--10797.

\bibitem[{Zhang, Robinson, and Tepper(2018)}]{DBLP:conf/esws/ZhangRT18}
Zhang, Z.; Robinson, D.; and Tepper, J.~A. 2018.
\newblock Detecting Hate Speech on Twitter Using a Convolution-GRU Based Deep
  Neural Network.
\newblock In \emph{ESWC}, 745--760.

\bibitem[{Zhou, Chen, and Yang(2021)}]{zhou2021multimodal}
Zhou, Y.; Chen, Z.; and Yang, H. 2021.
\newblock Multimodal learning for hateful memes detection.
\newblock In \emph{2021 IEEE International Conference on Multimedia \& Expo
  Workshops}, 1--6.

\bibitem[{Zhu(2020)}]{zhu2020enhance}
Zhu, R. 2020.
\newblock Enhance multimodal transformer with external label and in-domain
  pretrain: Hateful meme challenge winning solution.
\newblock \emph{arXiv preprint arXiv:2012.08290}.

\bibitem[{Zia, Castro, and Tyson(2021)}]{zia2021racist}
Zia, H.~B.; Castro, I.; and Tyson, G. 2021.
\newblock Racist or sexist meme? classifying memes beyond hateful.
\newblock In \emph{WOAH}, 215--219.

\end{thebibliography}

\end{document}